\PassOptionsToPackage{dvipsnames,table}{xcolor}

\documentclass{article}

\usepackage[final]{corl_2025}

\usepackage{amssymb}

\usepackage{booktabs}
\usepackage{multirow}
\usepackage{ulem}
\usepackage{enumitem}
\usepackage{booktabs} 

\usepackage[frozencache=true,cachedir=minted-cache]{minted}

\usepackage{graphicx}
\usepackage[most]{tcolorbox}
\tcbuselibrary{listings, skins, breakable}
\usepackage{algorithm}
\usepackage{algorithmic}
\usepackage{amsmath}
\usepackage{wrapfig}
\usepackage{xspace}

\usepackage{listings}

\usepackage{placeins}
\usepackage{listings}

\usepackage{etoc}

\usepackage[toc,page,header]{appendix}
\usepackage{minitoc}

\usepackage{tabularx}
\usepackage{geometry}

\usepackage{xcolor}
\usepackage{booktabs}

\newcommand{\name}{\textsc{prism}\xspace}

\newcommand{\namep}{\textsc{prism}'s\xspace}

\newcommand{\secvspace}{\vspace{-5pt}}

\title{Distilling On-device Language Models for Robot Planning with Minimal Human Intervention

\vspace{-2mm}
}

\author{
  \textbf{Zachary Ravichandran}, 
   \textbf{Ignacio Hounie},
   \textbf{Fernando Cladera},\\
   \textbf{Alejandro Ribeiro},
   \textbf{George J. Pappas},
  \textbf{Vijay Kumar} \\ \\
 University of Pennsylvania
 \vspace{-5mm}
}

\begin{document}
\maketitle


\begin{abstract}
Large language models (LLMs) provide robots with powerful contextual reasoning abilities and a natural human interface. 
Yet, current LLM-enabled robots typically depend on cloud-hosted models, limiting their usability in environments with unreliable communication infrastructure, such as outdoor or industrial settings.
We present \textsc{prism}, a framework for distilling small language model (SLM)-enabled robot planners that run on-device with minimal human supervision.
Starting from an existing LLM-enabled planner, \name automatically synthesizes diverse tasks and environments, 
elicits plans from the LLM, and uses this synthetic dataset to distill a compact SLM as a drop-in replacement of the source model.
We apply \name to three LLM-enabled planners for mapping and exploration, manipulation, and household assistance, and we demonstrate that \name improves the performance of Llama-3.2-3B from 10-20\% of GPT-4o's performance to over 93\% --- \textit{using only synthetic data}.
We further demonstrate that the distilled planners generalize across heterogeneous robotic platforms (ground and aerial) and diverse environments (indoor and outdoor).
We release all software, trained models, and datasets at \href{https://zacravichandran.github.io/PRISM}{https://zacravichandran.github.io/PRISM}.

\end{abstract}



%
\doparttoc 
\faketableofcontents

\section{Introduction}
\label{sec:intro}
\secvspace

Large language Models (LLMs) have emerged as powerful tools for robot planning by providing contextual reasoning and a natural human interface. 
LLM-enabled robots have been successfully deployed for  mobile manipulation~\cite{chen2022nlmapsaycan, momallm24, rana2023sayplan}, autonomous driving~\cite{omama2023altpilotautonomousnavigationlanguage, li2024llada}, service robots~\cite{llm_service_robot}, navigation~\cite{pmlr-v205-shah23b,pmlr-v229-shah23c}, and inspection~\cite{ravichandran_spine, fadhil2024saycomplyllm}.
Being trained on internet-scale data, LLMs 
contain rich semantic knowledge and are highly adaptable.
With only a few in-context examples, they can attune to task-relevant information and generalize to novel robotic scenarios.

However, the scale of LLMs brings significant computational demands. 
Current frontier models contain hundreds of billions of parameters and require significant cloud computing infrastructure~\cite{openai2023gpt4, grattafiori2024llama}. 
Although small language models (SLMs) can run on-device,
their relatively poor performance in robot planning domains has  precluded widespread adoption by the community~\cite{scalingupdistilling}.
Consequently, nearly all LLM-enabled robots rely on models such as GPT or PALM~\cite{dai2024optimalltlllm, chen2022nlmapsaycan, rana2023sayplan, ravichandran_spine, omama2023altpilotautonomousnavigationlanguage, liu2023llmp, song2023llmplanner, llm_service_robot}, which tethers them to external compute.
While this dependence may be feasible in structured environments such as homes or offices, it severely limits deployment in large-scale, unstructured settings --- domains critical for environmental monitoring, industrial inspection, disaster response, and a host of similar problems.

\begin{figure}[t!]
    \centering
    \includegraphics[width=1\linewidth]{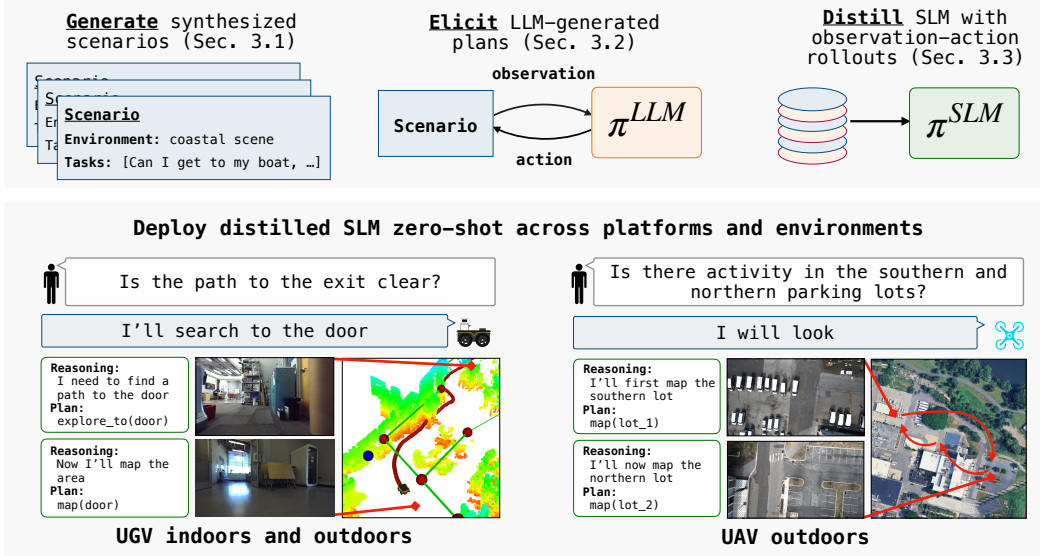}
    \vspace{-12pt}
    \caption{\textsc{PRISM} distills on-device language models for robot planning with minimal human intervention. Given an existing LLM-enabled planner, \name synthesizes diverse task and textual environments, elicits plans from the LLM, then uses the resulting dataset to fine-tune an SLM that rivals the performance of the original planner, while being extremely compute efficient. We demonstrate \textsc{prism}-distilled models across platforms and environments.
    \vspace{-8pt}}
    \label{fig:intro}
\end{figure}

To address these limitations, we introduce \textsc{prism} (Planning with Robotic dIstilled Small language Models), a framework for distilling SLMs that run on-device with minimal human intervention.
Starting from an existing LLM-enabled planner (termed ``source''), \name generates diverse tasks and textual environment representations --- such as scene graphs or object sets --- and uses these generations to elicit plans from the source LLM.
This interaction produces high-quality planning demonstrations without manual annotations, simulators, or dataset curation. The user simply provides the LLM-enabled planner along with its action and observation spaces.
\name then uses the generated dataset to fine-tune an SLM that shares the original planner's interface, enabling it to directly replace the source LLM.
We outline \name in Fig.~\ref{fig:intro}, we summarize \namep  \textbf{contributions}:

\begin{itemize}[left=1pt]
    \item \textbf{Effectiveness.} \name distills SLMs that rival the planning performance of a source LLM-enabled planner, such as GPT-4o, as measured by planning success rate.
    \item \textbf{On-device models.} \name effectively distills language models that are small enough to run onboard robot compute. In this work, we consider models with a memory footprint of under 5GB.
    \item \textbf{Minimal human interaction.} \name requires a high-level configuration comprising the source planner's observation space, action space, \textit{etc.}, but it does not require manual dataset construction, simulators, details about the deployment environment, or other labor intensive information.
\end{itemize}

We evaluate \name across three LLM-enabled planners --- SPINE on real-world air and ground robots~\cite{ravichandran_spine}, LLM-Planner in the ALFRED simulator~\cite{song2023llmplanner, ALFRED20}, and SayCan in a manipulation simulator~\cite{saycan2022arxiv}.
While undistilled SLMs achieve only 10-20\% of GPT-4o's success rate, \name boosts performance to over 93\% across all three settings.
Moreover, the distilled SLMs compute efficiency enables them to plan onboard a robot in closed-loop control settings.
We also release datasets, models, and code in order to promote reproducibility.
The rest of the paper is structured as follows. We overview related work in \S\ref{sec:citations}, describe our framework in \S\ref{sec:method}, present experimental results in \S\ref{sec:experiments},  provide concluding remarks in \S\ref{sec:conclusion}, and discuss limitations in \S\ref{sec:limitations}.

\secvspace


\section{Related work}
\label{sec:citations}
\secvspace

\textbf{LLM-Enabled Robots.} LLMs have emerged as a powerful enabler of contextually-aware robots, as evidenced by their deployment in mobile manipulation~\cite{chen2022nlmapsaycan, rana2023sayplan, momallm24, saycan2022arxiv, codeaspolicies2022}, service robots~\cite{llm_service_robot}, task and motion planning~\cite{chen2023autotamp}, autonomous driving~\cite{sharan2023llm, omama2023altpilotautonomousnavigationlanguage}, anomaly detection~\cite{tagliabue2023real, SinhaElhafsiEtAl2024Aesop}, navigation~\cite{dai2024optimalltlllm, pmlr-v229-shah23c, pmlr-v205-shah23b, xie2023reasoning}, exploration and mapping~\cite{ravichandran_spine, fadhil2024saycomplyllm}, and multi-robot teaming~\cite{llm_team}.
These works forego any model training; instead, they configure a pre-trained 
LLM via in-context examples.
While convenient and effective, this approach requires the use of extremely large models (over 100B parameters) ---  \emph{all} the above works use GPT-4~\cite{openai2023gpt4},  PALM~\cite{chowdhery2022palm}, or similar --- which are far too large to run onboard a robot and therefore tethers the robot to high-quality network infrastructure.

\textbf{Distilling LLMs.} LLM distillation aims to retain the contextual reasoning abilities of LLMs while overcoming their compute limitations.
These methods train an SLM to emulate the generations of a source LLM via supervised fine tuning (SFT)~\cite{alpaca, vicuna2023, lamini-lm}, reinforcement learning~\cite{bai2022constitutional, cui2023ultrafeedback}, and ranking optimization~\cite{rafailov2023direct,song2024preference}, among other methods.
By finetuning a target SLM rather than training from scratch, these methods enjoy significant compute and data efficiency~\cite{hu2022lora}.
While traditional distillation methods curate a dataset of queries with which to elicit LLM generations, an emerging line of \textit{self-distillation} research uses an LLM to generate \textit{both} queries and responses~\cite{liu2024evolving, baykal2022robust, wang2022self},
thereby reducing the need for dataset curation.
In contrast to prior self-distillation methods which operate in the virtual domain, methods for robot planning must be \textit{grounded} in the physical world by respecting robot affordances, responding to sensory input, and operating in closed-loop settings. 
While our methodology draws from general distillation techniques, its key novelty lies in addressing the unique challenges of robot planning, which prior distillation work has not considered.

\textbf{Distilling Robot Planners}. 
Distillation has been applied to virtual planning domains, including web search or video games~\cite{chen2023fireact, zeng2023agenttuning, yin2023lumos, kong2023tptuv2boostingtaskplanning, choi2024embodied}. 
These methods typically use high-quality queries --- such as a dataset of instructions --- to elicit planning examples from an LLM, which are then used to train a smaller model.
\citet{autoact} apply self-distillation principles to this realm by bootstrapping data with generations from an LLM.
Recently, roboticist have applied distillation techniques to problems including multi-robot control~\cite{liu2024languagedrivenpolicydistillationcooperative} and manipulation~\cite{llmknowledgedistillation}.
Of particular note is the work by \citet{scalingupdistilling}, who use an LLM to generate expert plans which are used to distill a visuomotor policy. 
While these methods successfully leverage the contextual knowledge of an LLM to train downstream planners, 
they still require external knowledge ---  such as a simulator or curated data --- and these works distill auxiliary planners that cannot replace the source LLM.
In contrast, \name \textit{synthesizes all training data} directly from a source LLM-enabled planner.
The resulting distilled SLM is not an auxiliary component but a compact planner designed to \textit{directly replace the source LLM,} making it deployable onboard robots without network reliance.

\secvspace

\section{Distilling On-device LMs for Robot Planning}
\label{sec:method}
\secvspace

\name distills an SLM that mimics the performance of an LLM-enabled planner with minimal human intervention --- \textit{i.e.}, without reliance on datasets or simulators. 
Our key observation is that because the action and observation spaces of an LLM-enabled planner are textual, 
they can be synthesized and used to elicit plans from the LLM, and the resulting data can be used to train an SLM.

Concretely, \name takes as input a \textit{source} LLM-enabled planner, $\pi^{\text{LLM}}(a_i | o_i, s)$, which is conditioned on a task $t$, takes an observation $o_i$, and produces an action $a_i$, at planning iteration $i$. These input-output sets must be textually represented, but we place no further restrictions on their format.
\namep objective is to train a \textit{target} SLM --- \textit{e.g.,} Llama-3.2-3B or similar --- to produce a planner, $\pi^{\text{SLM}}(a_i | o_i, t)$ that mimics the source LLM while being significantly more efficient.

\name comprises three modules --- \textit{scenario generation}, \textit{plan elicitation}, and \textit{planner distillation}.
The scenario generation module synthesizes a diverse set of tasks and environments (termed ``scenarios,'' \S\ref{sec:scenario_generation}).
These scenarios are used by the elicitation module to extract plans from the LLM-enabled planner, and this process yields a dataset of task-observation-action sequences (\S\ref{sec:plan_elicit}).
Finally, the planner distillation module uses this synthesized dataset to train a target SLM via SFT (\S\ref{sec:distillation}). 
We discuss each module in detail below.

\secvspace

\subsection{Scenario Generation}
\label{sec:scenario_generation}
\secvspace
The scenario generation module takes as input the LLM-enabled planner's action and observation spaces and produces \textit{scenarios} comprising environments --- which produce observations for the planner --- and semantically coherent tasks.
We instantiate this module with an LLM (GPT-4o), given their generative abilities. 
This scenario generation LLM is prompted with the observation format along with the desired environment size and number of tasks, which are hyperparameters of this module. 
This generation procedure, combined with a high temperature (1 in our work), promotes data diversity. We provide the high-level format; please see the appendix for the full prompt:
\begin{tcolorbox}[colback=gray!3, colframe=black, left=1mm, right=1.5mm, top=1.5mm, bottom=1mm] \small
\begin{minted}{text}
You are generating data for training an llm-enabled planner.
The planner takes observations in the following format: <OBSERVATION>.
Generate an environment of size <SIZE> with <N_TASKS> corresponding tasks
and a description of the scenario.
\end{minted}
\end{tcolorbox}
Along with \texttt{environments} and \texttt{tasks}, the generator provides a \texttt{description} to enforce semantic coherence.
All generations are verified to ensure environments adhere to the representation required by the source planner; the generator is re-prompted if it provides an erroneous environment.

In our instantiation of \textsc{prism}, we consider two observation representations.  First, we consider a scene graph with nodes of type \texttt{region} or \texttt{object}; \texttt{object} nodes represent semantic entities, while \texttt{region} nodes are traversable points in the scene. 
Edges in the graph are defined between two regions (``region nodes'') or an object and a region (``object nodes''). 
Region edges indicate traversable paths, while object edges indicate visibility from a particular location. 
Nodes can also be enriched with semantic descriptions derived from a vision-language model or similar perception system.
We next consider object sets --- a popular representation for manipulation and instruction-following tasks --- which define all objects accessible by the planner.
If using the scene graph representation, an output from the scenario generation LLM would look like the following:

\vspace{3pt}

\begin{minipage}{0.5\textwidth}
\begin{tcolorbox}[colback=gray!3, colframe=black, left=1mm, right=1.5mm, top=1.5mm, bottom=1mm, width=1\textwidth,
title=Scenario generation format,
coltitle=black,        
colbacktitle=white,    
fonttitle=\bfseries,
boxrule=0.5pt,
boxsep=1mm
] \small
\begin{minted}{json}
{"description": "scene description",
"environment": "environment",
"task": ["task_1", "task_2", "..."]}
\end{minted}
\end{tcolorbox}
\end{minipage}
\begin{minipage}{0.45\textwidth}
\begin{tcolorbox}[colback=gray!3, colframe=black, left=1mm, right=1.5mm, top=1.5mm, bottom=1mm, width=1\textwidth,
title=Graphical environment representation,
coltitle=black,        
colbacktitle=white,    
fonttitle=\bfseries,
boxrule=0.5pt,
boxsep=1mm
] \small
\begin{minted}{json}
{"regions": ["..."],
"objects": ["..."],
"edges": ["..."]}
\end{minted}
\end{tcolorbox}
\end{minipage}

\begin{figure}[t!]
    \includegraphics[width=1\textwidth]{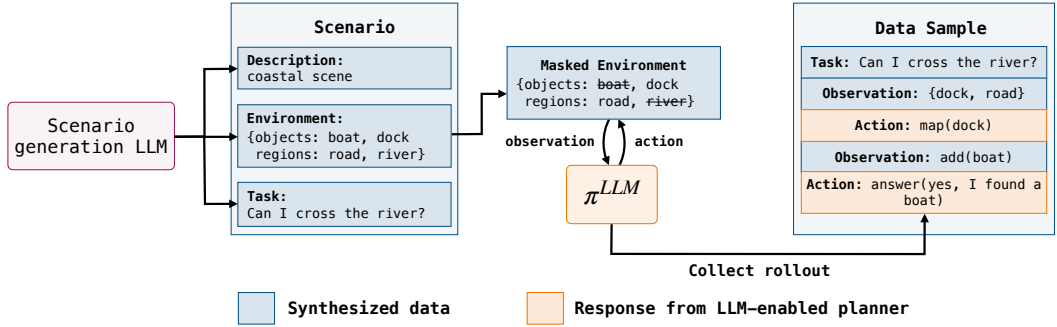}
    \vspace{-12pt}
    \caption{Synthesizing a  data sample: The scenario generator synthesizes an environment and semantically coherent task. The environment is masked, yielding an initial partial observation for the LLM-enabled planner. Given this task and partial observation, the LLM-enabled planner iteratively provides actions and receives new observations until the task is complete. This rollout is collected as a data sample and added to the synthesized dataset.
    \vspace{-8pt}}
    \label{fig:data-gen}
\end{figure}

\subsection{Plan Elicitation}
\label{sec:plan_elicit}
\secvspace

Given the generated scenarios, the plan elicitation module extracts plans from the source LLM-enabled planner via three steps  --- \textit{environment masking}, \textit{closed-loop planning emulation}, and \textit{plan validation}.
Environment masking produces a partial observation, $o^1$, for the planner by hiding a randomized portion of the environment --- \textit{i.e.,} removing nodes and edges from a graph, removing objects from a set, \textit{etc.}.
The plan elicitation module then emulates closed-loop planning by using the provided action space.
Each action defines some operation on the environment --- for example,  a \texttt{map} action may reveal previously unobserved objects, a \texttt{goto} action would change the robots location, \textit{etc.}
The plan elicitation module uses this action-observation relationship to simulate what the planner \textit{would have seen} had it taken a certain action in a synthesized environment --- \textit{i.e.,} observation $o^{i+1}$ given action $a^i$ and the synthesized environment.
In our instantiation of \name, we consider actions for mapping, navigation, and object interaction, though more actions may be considered if needed.
Given this ability to emulate closed-loop planning, the module seeds the LLM-enabled planner with the initial observation and iterates until termination.
Termination occurs when the planner signals task completion (\textit{e.g.,} via a \texttt{done} action) or a reaches a pre-defined maximum number of planning iterations.
Should the planner timeout, we consider the sample \textit{invalid}. 
This simple validation procedure has significant effects on distilled SLM performance (see \S\ref{sec:ablate}), and this process yields a rollout of observation-pairs that is collected into dataset:
\begin{align}
    \mathcal{D} = \{s_i, (o_i^1, a_i^1, \dots, o_i^T, a_i^T)\}_{i=1}^N
\end{align}
where $N$ is the number of scenarios, and $T$ is the plan trajectory length (which may differ across scenarios). We 
provide an example of plan elicitation procedure in Fig.~\ref{fig:data-gen}.

\secvspace

\subsection{Planner Distillation}
\secvspace
\label{sec:distillation}
Given the dataset $\mathcal{D}$, the planner distillation module trains the target SLM-enabled planner $\pi^{\text{SLM}}$ using supervised fine-tuning (SFT), minimizing the cross entropy  over every action $a_i^1 \ldots a_i^T$: 

\begin{align}\label{eq:objective}
   \min_{\pi^{SLM}} \frac{1}{N}\sum_{i=1}^N \sum_{t=1}^T CE\left(\pi^{SLM}\left(a_i^t | s_i, o_i^1, a_i^1, \dots o_i^{t-1}. a_i^{t-1}\right), \; a_i^t\right)
\end{align}

Each action prediction utilizes the cumulative context of all preceding observation-action pairs. 
We train on all actions jointly, which has shown improved efficiency without performance degradation compared to more elaborate training schemes in multi-turn SFT of LMs~\cite{sft-multiturn}. Since the objective in Eq.~\ref{eq:objective} only involves actions from the LLM planner, it can be evaluated in a single forward-pass without the overhead of autoregressive sampling of SLM actions. 
To reduce GPU memory footprint during finetuning, we leverage LoRA~\cite{hu2022lora}, a popular parameter efficient fine tuning technique. LoRA  adds a low rank trainable update to pre-trained model weights,  
thereby shrinking optimizer states. For a 3 billion parameter Llama, LoRA with rank 8 requires only 6 million trainable parameters. Since updates can be merged with the weights, it incurs no additional inference cost.
And while \namep primary objective is to produce a small language model for robot planning, employing efficient optimization techniques is a further advantage of our method. 

\secvspace
\section{Experimental Results}
\label{sec:experiments}
\secvspace

We evaluate \name in order to assess how well it fulfills the contributions outlined in \S\ref{sec:intro}. 
By design, \name requires \textbf{minimal human intervention}, 
and we empirically evaluate the second two contributions.
First, we evaluate \textbf{effectiveness}: does \name distill SLM-enabled planners with comparable performance to the source LLM-enabled planner? 
We then evaluate \textbf{efficiency}:  do the produced models run on-device in real-time?
We also measure the impact of two key components of our data synthesis pipeline --- environment masking and plan validation.
We evaluate \name over three LLM-enabled planners~\cite{ravichandran_spine,saycan2022arxiv,song2023llmplanner} in distinct environments and tasks with the following methodology.
We first evaluate the planner using an LLM, as it was original designed.
We then replace the planner's LLM with an SLM, and we measure performance.
Finally, we apply \name to the planner, and we measure the performance of the planner while using the \name-distilled SLM.

\begin{figure}[h!]
    \centering
    \includegraphics[width=0.99\linewidth]{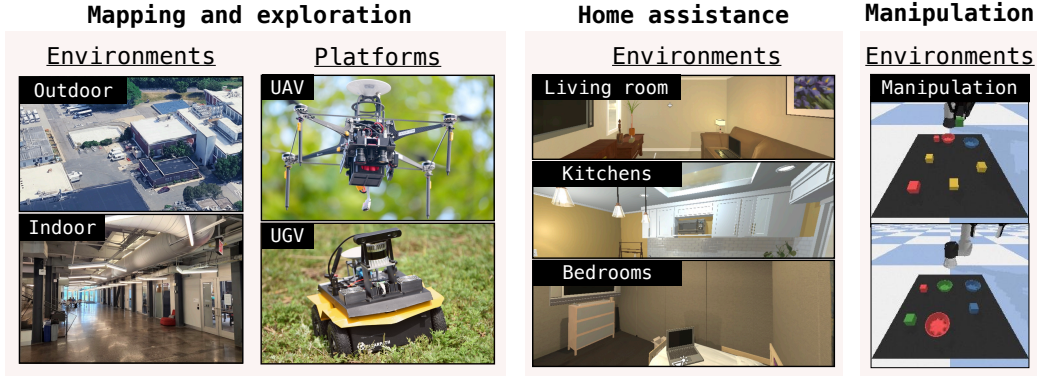}
    \caption{We evaluate \textsc{PRISM} on three distinct LLM-enabled planning domains: mapping and exploration, home assistance, and manipulation. \uline{Left:} SPINE~\cite{ravichandran_spine} in both indoor and outdoor environments on bot UAVs and UGVs. \uline{Middle:} LLM-Planner~\cite{song2023llmplanner} to household tasks in the ALFRED~\cite{ALFRED20} simulator. \uline{Right:} SayCan~\cite{saycan2022arxiv} to manipulation tasks. }
    \vspace{-8pt}
    \label{fig:exp-setup}
\end{figure}

\secvspace
\subsection{Experimental setup}
\label{sec:planner_setup}
\secvspace
We consider the following three LLM-enabled planners.
For each planner, we describe the evaluation tasks and experimental environments, as outlined in Fig.~\ref{fig:exp-setup}.

\textbf{SPINE}~\cite{ravichandran_spine}  is designed for language-driven navigation, mapping and exploration in partially-known and unstructured environments.
The planner operates in a closed-loop manner: actively mapping its scene and updating its plan based on its findings. 

\begin{itemize}[left=5pt]
\item \textbf{Tasks.}
We consider two levels of difficulty.
First, we evaluate against \textit{fully-specified} tasks that directly reference actions and semantics in the scene (\textit{i.e.,} ``inspect the desk'').
Next, we evaluate against \textit{under-specified} tasks that require the planner to infer relevant actions and semantics (\textit{i.e.,} ``find me a place to work'').
We consider 20 indoor UGV tasks, 15 outdoor UGV tasks, and 10 UAV tasks, split evenly between fully- and under-specified variants.

\item \textbf{Environments.}
We evaluate SPINE in both indoor and outdoor environments with the following levels of difficulty: \textit{mapping} tasks require the robot to collect semantic information, and  \textit{exploration} tasks require the robot to collect both metric and semantic information --- \textit{i.e,} explore free space. We consider 15 exploration tasks and 30 mapping tasks.

\end{itemize}

\textbf{SayCan}~\cite{chen2022nlmapsaycan} is a  hierarchical language‑guided planner that combines an LLM’s task‑level reasoning with lower‑level affordance estimates. Instead of token-wise probabilities used in the original setting, we prompt the LLM to generate candidate actions consistent with our framework’s textual input–output planner abstraction.
\begin{itemize}[left=5pt]
    \item \textbf{Tasks.}  A robot arm is tasked with rearranging colored bowls and boxes on a table. In the absence of an open-source benchmark, we design twenty tasks, each rendered in multiple scene variants. Following~\cite{ren2023robotsknowno}, we introduce \textit{Attribute ambiguity} --- replacing canonical terms with synonyms --- and \textit{Numeric ambiguity}  --- swapping exact counts for vague phrases such as a  ``some.'' We also include more challenging variants with arithmetic constraints, color--based negation, and load balancing, forcing the planner to reason over these specifications. See Appendix for details.
    \item \textbf{Environments}. We use the 
PyBullet~\cite{coumans2016pybullet} simulator. For each task, we create multiple environments with different configurations of colored blocks and bowls, and randomized placings.
\end{itemize}

\textbf{LLM-Planner}~\cite{song2023llmplanner}  casts language-driven household manipulation tasks as few-shot text-to-plan generation. The planner queries an LLM (GPT4) with the task specification, an object-set description of the observed environment, and 9 in-context examples; these in-context examples are chosen by matching LLM embeddings between the input task and a pre-defined set of 100 expert demonstrations. 
The LLM returns a sequence of semantic sub-goals (e.g., \texttt{Navigation fridge}, \texttt{OpenObject fridge}) which, if feasible, are realized by a lower-level controller

\begin{itemize}[left=5pt]
\item \textbf{Tasks.} We consider tasks from Action Learning From Realistic Environments and Directives (ALFRED)~\cite{ALFRED20}. We evaluate on the seven ALFRED templates: \textit{Pick \& Place}, \textit{Stack \& Place}, \textit{Pick-Two \& Place},  \textit{Clean \& Place}, \textit{Heat \& Place}, \textit{Cool \& Place}, and \textit{Examine-in-Light}, and we report results in the Validation split. Directives provide only a single goal sentence, so the planner must infer all concrete actions.

\item \textbf{Environments.} 
Experiments are run in the \textit{unseen} split of ALFRED -- i.e., environments differ from the in-context examples from the training set -- comprising eight AI2-THOR~\cite{kolve2017ai2thor} homes with kitchens, bedrooms, bathrooms, and living rooms. The agent must explore cluttered, partially observable rooms containing up to 115 object types.
\end{itemize}

\textbf{Baselines and training:} We compare \name against the above planners equipped with the following language models: GPT-4o (\textbf{LLM}) and Llama-3.2-3B (\textbf{SLM}), which we choose because of their widespread use and competitive performance within their size class; we defer comparative studies to future work. Moreover, the  selected models have acceptable latency for the experimental setup mentioned above.
Together, these baselines allow us to measure both the comparative improvement \name provides over a base SLM relative to a SoTA LLM.
\namep trains on 3,200 examples for SPINE, 1,500 examples for SayCan, and 1000 examples for LLM-Planner.

\begin{wraptable}{r}{0.47\textwidth}
\centering
\scriptsize
\begin{tabular}{l|ccc}
\toprule
 & \multicolumn{3}{c}{\textbf{Success rate (\%)}}\\
\textbf{Method} & SPINE~\cite{ravichandran_spine} & SayCan~\cite{chen2022nlmapsaycan} & LLM-Planner~\cite{song2023llmplanner} \\
\midrule
SLM  & 13.54  & 10.0  & 1.76   \\
\name & 92.00 & 50.0 &  15.54   \\
LLM & 98.07 & 50.0 & 16.49  \\
\bottomrule
\end{tabular}
\caption{Comparison of SLM, \name, and GPT-4o in SPINE~\cite{ravichandran_roboguard}, SayCan~\cite{saycan2022arxiv}, and LLM-Planner~\cite{song2023llmplanner}}
\label{tab:main}
\end{wraptable}

\secvspace
\vspace{3pt}
\textbf{Hardware Platforms.}
We evaluate SPINE on two physical platforms. First, we consider a UGV --- a Clearpath Jackal ---  equipped with an Ouster OS1-64 LiDAR and ZED-2i for object detection and obstacle avoidance, and compute comprising a Ryzen 5 3600 CPU and an Nvidia RTX 4000 SSF Ada GPU.
Second, we consider the Falcon 4 UAV, equipped with a downward-facing RGB camera and an Nvidia Jetson Orin NX computer.

\subsection{Model Effectiveness}
\secvspace
\begin{figure}[t]
    \centering
    \includegraphics[width=1\textwidth]{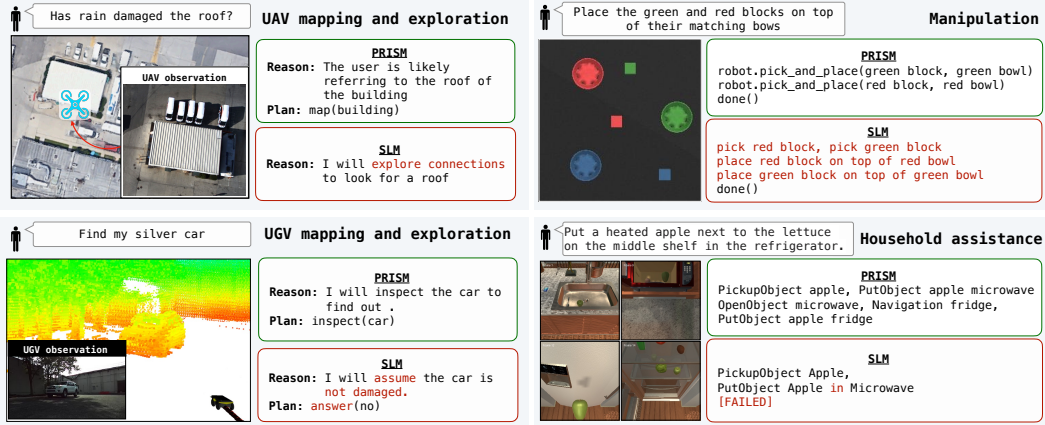}
    \vspace{-12pt}
    \caption{In the following examples, \name completes the task, and we highlight failure modes of the SLM. \uline{Top right:} SLM does not identify semantic targets. \uline{Top left:} SLM incorrectly invokes planner API and mistakes logical ordering; two blocks cannot be simultaneously help by the robot. \uline{Bottom right:} SLM makes unwarrented assumptions about the environment. \uline{Bottom left:} SLM incorrectly invokes API and the plan terminates.}
    \vspace{-12pt}
    \label{fig:compare_planner}
\end{figure}

\begin{wraptable}{r}{0.5\textwidth}
\centering
\scriptsize
\begin{tabular}{cccc|cccc}
\toprule
 & \multicolumn{3}{c}{Fully-specified (\%)} & \multicolumn{3}{c}{Under-specified (\%)} \\
\cmidrule(lr){2-4} \cmidrule(lr){5-7}
 & LLM & Ours & SLM & LLM & Ours & SLM \\
\midrule
Mapping & 100 & 100 &  50 & 100 & 90 & 0 \\
Exploration & 100 & 87.5 & 0 & 83.3 & 75 & 0 \\
\bottomrule
\end{tabular}
\caption{Planning success rate in varying levels of environment and task difficulty}
\label{tab:difficulty}
\vspace{-4pt}
\end{wraptable}
We report success across all three planners in Table~\ref{tab:main}.
GPT-4o achieves a 98.07\%, 50\%, and 16.49\% success rate across SPINE, SayCan, and LLM-Planner respectively, which match trends from the original publications (SayCan is slightly lower due to our custom benchmark).
Without distillation, SLM achieves 13.54\%, 10\%, and 1.76\% success.
\name boosts this performance to 92\%, 50\%, and 15.54\%, which amounts to at least a 93\% increase across all three planners.
Qualitatively, \name addresses two primary failure modes of the SLM. 
First, the SLM will provide a feasible plan (\textit{i.e.,} using the correct syntax), but the plan will be semantically incorrect.
In the second failure mode, the SLM fails to provide a valid plan altogether. 
We highlight comparative examples in Fig.~\ref{fig:compare_planner}.

We then analyze performance over varying levels of task and environment difficulty in SPINE, as reported in Tab.~\ref{tab:difficulty}.
In the easiest scenarios --- fully specified mapping tasks --- GPT-4o and \name both achieve 100\% success rate, while SLM only achieves 50\% success.
We observe a modest (10\%) performance decrease for \name in under-specified tasks.
Exploration presents the most difficult setting for all models.  
\namep performance drops to 75\% in under-specified exploration, while GPT-4o only achieved 83.3\% success.
Notably,  SLM performance drops to 0\% in difficulty settings beyond fully-specified mapping missions.
These results suggest that spatial reasoning is still comparatively difficult for LLM-enabled planners, and \name inherits this weakness.

\secvspace

\subsection{Model efficiency}
\secvspace
\begin{wrapfigure}{r}{0.45\textwidth}
    \includegraphics[width=1\linewidth]{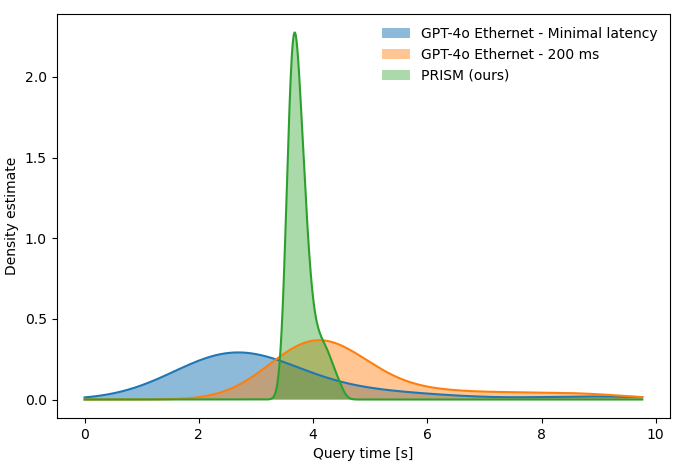}
    \vspace{-12pt}
    \caption{Latency comparison between GPT-4o and \name}
    \label{fig:latency}
\end{wrapfigure}
We next perform latency analysis between \name-distilled models and GPT-4o --- our baseline LLM --- using SPINE running on the UGV (see \S\ref{sec:planner_setup}).
We measure the latency of 60 \name queries, and we compare this to 60 GPT-4o queries under the following network simulations.
\textit{Ethernet - Minimal latency} corresponds to the best scenario for querying GPT-4o, where the robot is connected to the internet using an ethernet connection. The round-trip time to \texttt{openai.com} is 6.7 ms, and we obtain a mean query time of 3.64 s, with a standard deviation of 2.25 s.
\textit{Ethernet - 200 ms} adds 100ms of latency to each received and transmitted packet using the utility traffic controller for Linux~\cite{networking:iproute2_2023},
which simulates the conditions of a robot using satellite internet services and mesh networks. 
In this setting, we obtain a mean query time of 4.90~s, with a standard deviation of 1.73~s.
Fig.~\ref{fig:latency} shows a density estimate for the different evaluated methods. We observe that the latency of \name is within 200 ms of GPT-4o under ideal conditions. \name  is more than 1 second faster than GPT-4o under realistic deployment conditions, while exhibiting less variance. This highlights a major advantage of \name: besides being able to operate \textit{without any network}, it offers more \textit{deterministic} latency as it does not depend on uncertainties in the network conditions.

\secvspace

\subsection{Importance of environment masking and plan validation}
\secvspace
\label{sec:ablate}
\begin{wraptable}{r}{0.3\textwidth}
    \centering
    \scriptsize
    \begin{tabular}{c ccc}
    \toprule
     Method & Success\\
     \toprule
     Ours & 90.9\%\\
     (-) environment masking & 63.6\%\\
     (-)  validation &  45.5\% \\
        \bottomrule
    \end{tabular}
    \caption{Data synthesis ablation}
    \label{tab:data_gen}
\end{wraptable}
Finally, we measure the importance of two key components of our data synthesis framework ---  environment masking and plan validation ---  by ablating these components.
We evaluate performance across eleven tasks and environments sampled from the larger collection described in \S\ref{sec:planner_setup}, and we report result in Tab.~\ref{fig:data-gen}.
Without environment masking, performance decrease by 27.3\% to 63.6\%. Having not been trained with interactive data, the this variant fails when success require many planning iterations.
Without validation, performance drops by nearly half to 45.5\%.
The no-validation variant mimics failure modes from source LLM not properly reacting to new observations and repeatedly calling the same, ineffective action; these failure modes are typically filtered out with validation.

\secvspace


\section{Conclusion}
\label{sec:conclusion}
\secvspace

We present \textsc{prism}, a framework for distilling effective SLM-enabled robot planners with minimal human interaction.
Given a source LLM-enabled planner, \name synthesizes tasks and environments, elicits plans from the LLM-enabled planner in these synthesized environments, and then uses the resulting data to train an SLM-enabled planner that serves as a drop-in replacement for the source model.
We evaluate \namep ability to distill SLMs that rival the performance of  LLMs on three distinct planning domains. 
We find that while an undistilled SLM (Llama-3.2-3B) achieves between 10-20\% performance of GPT-4o, \name boosts this to over 93\% \textit{using only synthetic data}, and we successfully deploy \name-distilled models for planning on air and ground robots.
Future work may investigate methods for synthesizing higher quality data --- in particular, data that leads to improved spatial reasoning of the distilled SLMs.

\section{Limitations}
\label{sec:limitations}	

We identify two primary limitations to our work.
Firstly, we consider LLM-enabled planners that operate over atomic action primitives, such as those for navigation, mapping, and manipulation.
While such action spaces encompass a large portion of the existing literature~\cite{rana2023sayplan, ravichandran_spine, fadhil2024saycomplyllm, saycan2022arxiv, vlmaps, momallm24, pmlr-v229-shah23c}, a class of LLM-enabled planners use formal languages, code, or other action representations~\cite{codeaspolicies2022,liu23lang2ltl,chen2023nl2tl}.
Because the latter action representations are still textual, \name is likely comparable with these methods,
though future work should validate this hypothesis. 
Secondly, recent work has shown that even frontier LLMs can produce harmful robotic actions~\cite{robey2024jailbreaking}.
We anticipate that \name inherits these safety vulnerabilities, and we
 propose two mitigation strategies.
In the short term, our work is compatible with recently-proposed safeguards for LLM-enabled robots~\cite{ravichandran_roboguard, verify_llm_ltl, yang2023plug}. 
Future research could also directly incorporate safety objectives into \namep distillation process~\cite{bai2022constitutional,ouyang2022traininglanguagemodelsfollow}.


\acknowledgments{
We gratefully acknowledge support from the Distributed and
Collaborative Intelligent Systems and Technology (DCIST),
NSF Grant CCR-2112665, DARPA SAFRON grant HR0011-25-3-0135, and the NSF Graduate Research Fellowship.
We thank Antonio Loquercio and Dinesh Jayaraman for their support during the inception of this work in \textit{ESE 6800: Real World Robot Learning}, and Varun Murali and Anish Bhattacharya for their helpful feedback on the manuscript.}

\bibliography{main}

\newpage

\appendix

\section{Summary of Appendices}

We report the implementation details of our LLM distillation procedure in \S\ref{sec:appendix_dist}.
We then provide further detail on the action primitives, output format, observation space, and evaluation tasks for the three planners used in our evaluation --- SPINE (\S\ref{sec:appendix_spine}), SayCan (\S\ref{sec:appendix_saycan}) and LLM-Planner (\S\ref{sec:appendix_llm_planner}).
We provide the prompts used for the scenario generation LLM as applied to each planner (\S\ref{sec:appendix_scenario_gen}).
Finally, we provide examples of synthesized environments and elicited plans for each planner (\S\ref{sec:appendix_examples}).

\section{LLM Distillation Implementation}
\label{sec:appendix_dist}

We use the Unsloth\footnote{https://unsloth.ai/} Library, which is built on top of the Huggingface Transformers library and PyTorch.
We report hyperparameters used for distilling SLMs via supervised fine tuning in Tab.~\ref{tab:slm_hyperparameters}

\begin{table}[h!]
    \centering
    \begin{tabular}{c|c|c|c}
        \toprule
        Parameter & SPINE  & SayCan & LLM Planner\\
        \midrule
        learning rate & 1e-4 & 1e-4 & 2e-4\\
        weight decay & 1e-3 & 1e-3 & 5e-2\\
        epochs & 5 & 5 & 5 \\
        warmup steps & 5 & 5 & 5 \\
        lora rank & 32 & 32 & 16 \\
        lora alpha & 16 & 16  & 16 \\
        \bottomrule
    \end{tabular}
    \caption{SLM hyperparameters}
    \label{tab:slm_hyperparameters}
\end{table}

\section{SPINE}
\label{sec:appendix_spine}

SPINE~\cite{ravichandran_spine}: The SPINE planner is designed for language-driven mapping, navigation, and exploration in partially-known and unstructured environments.
The planner operates in a closed-loop manner: actively mapping its scene and updating its plan based on its findings. 

\subsection{Action primitives} SPINE composes plans with the following action primitives.    
\begin{enumerate}[noitemsep, topsep=0pt, left=12pt]
    \item \texttt{map\_region(region\_node)}: Go to a region and search or objects.
    \item \texttt{inspect(object\_node, query)}: Inspect an object for attributes described in the query.
    \item \texttt{explore\_region(region\_node, exploration\_radius)}: Go to a region and explore within a given radius.
    \item \texttt{goto(region\_node)}: goto a region node.
    \item \texttt{answer(message\_to\_user)}: Terminate task and provide a message to the user.
\end{enumerate}

\subsection{Output format}

At each iteration, the LLM provides its primary goal (\texttt{primary\_goal}) --- which is particularly relevant for under-specified tasks --- relevant portions of the semantic graph for achieving that goal (\texttt{relevant\_graph}), its \texttt{reasoning} for its plan, and the \texttt{plan}, all via JSON

\begin{tcolorbox}[colback=gray!3, colframe=black, left=1mm, right=1.5mm, top=1.5mm, bottom=1mm, width=1\textwidth,
coltitle=black,        
colbacktitle=white,    
fonttitle=\bfseries,
boxrule=0.5pt,
boxsep=1mm
] \small
\begin{minted}{json}
{"primary_goal": "",
"relevant_graph": "",
"reasoning": "",
"plan": ""}
\end{minted}
\end{tcolorbox}

\subsection{Observation space}

The observation space is a semantic graph with nodes of type \texttt{regions} and \texttt{objects} and edges defined between and object and region (\texttt{object\_edges}) or two regions (\texttt{region\_edges}).
The initial observation is provided to the LLM via the JSON format below.

\begin{tcolorbox}[colback=gray!3, colframe=black, left=1mm, right=1.5mm, top=1.5mm, bottom=1mm, width=1\textwidth,
coltitle=black,        
colbacktitle=white,    
fonttitle=\bfseries,
boxrule=0.5pt,
boxsep=1mm
] \small
\begin{minted}{json}
{"regions": "",
"objects": "",
"region_connections": "",
"object_connections": ""}
\end{minted}
\end{tcolorbox}

During an episode, the planner receives updated observations via the graph-manipulation API: \texttt{add\_nodes}, \texttt{remove\_node}, \texttt{add\_connections}, \texttt{remove\_connections}, \texttt{update\_robot\_location}, 
\texttt{update\_node\_attributes}.

\subsection{Tasks}

Experimental tasks for the UAV and UGV are reported in Tab.~\ref{tab:ugv_task_indoor}, Tab.~\ref{tab:ugv_task_outdoor}, and Tab.~\ref{tab:uav_task_outdoor}.

\renewcommand{\arraystretch}{1.3}
\rowcolors{2}{gray!10}{white}

\begin{table}[h!]
\begin{tabularx}{\textwidth}{>{\raggedright\arraybackslash}X >{\raggedright\arraybackslash}X}
\toprule
\textbf{Fully-specified Indoor}  & \textbf{Under-specified Indoor} \\
\midrule
Go to the desk & Is the path to the exit clear? \\
Go to the table & Find a place to sit \\
Inspect the chair & Find a place to work \\
Map the area near the desk & I want to watch a movie. Help \\
Go to the door & I need to do some robotics experiments \\
Go to the ground\_1 & Go to the southern part of the scene \\
Inspect the television & Help me find my backpack. I last saw it when I was working. \\
Go inspect the table then return to where you started & I lost my laptop. Help. \\
Inspect the plant then return to where you started & What is near the person? \\
Inspect the chair then inspect the table & I will be giving a tour of the lab for kids. Is there anything dangerous around? \\
\bottomrule
\end{tabularx}
\caption{Indoor UGV tasks}
\label{tab:ugv_task_indoor}
\end{table}

\begin{table}[h!]
\begin{tabularx}{\textwidth}{>{\raggedright\arraybackslash}X >{\raggedright\arraybackslash}X}
\toprule
\textbf{Fully-specified Outdoor} & \textbf{Under-specified Outdoor} \\
\midrule
Map the path  & Where can I park my car? \\
Go to the trees  & I want a shady place to read   \\\
Inspect the car   & Are the chairs under the trees occupied?   \\
Go to sidewalk 2  & Is my car damaged?   \\
Go to parking lot 1 &  Is there debris on the path?  \\
&Is construction blocking the sidewalk?  \\
&I lost my car. Help.  \\
&What is near the person? \\ 
&Is the path clear?  \\
&Can I get to the parking lot?
\\
\bottomrule
\end{tabularx}
\caption{Outdoor UGV tasks}
\label{tab:ugv_task_outdoor}
\end{table}

\renewcommand{\arraystretch}{1.3}
\rowcolors{2}{gray!10}{white}

\begin{table}[h!]
\begin{tabularx}{\textwidth}{>{\raggedright\arraybackslash}X >{\raggedright\arraybackslash}X}
\toprule
\textbf{Fully-specified} & \textbf{Under-specified} \\
\midrule
Go to parking lot 1  & Is there space to park my car? \\
Map the building  & Is construction blocking the road? \\
Go to the canine center  & Find my dog \\
Go to the landing strip & Has rain damaged any roofs? \\
Map the bike rack & Is the playground built? \\
Return to the landing strip & Return to land  \\
\bottomrule
\end{tabularx}
\caption{UAV tasks}
\label{tab:uav_task_outdoor}
\end{table}

\section{SayCan}
\label{sec:appendix_saycan}

SayCan~\cite{chen2022nlmapsaycan} is a hierarchical language‑guided planner that combines an LLM’s task‑level reasoning with lower‑level affordance estimates. We  follow the tabletop rearrangement environment. Lower level language skill primitives are implemented with CLIPort~\cite{shridhar2022cliport}.

\subsection{Action primitives}
Actions are specified as \texttt{robot.pick\_and\_place(object, place)}. Objects and places must be listed in the observation. For example, ``\texttt{robot.pick\_and\_place(red block, blue bowl)}''.

\subsection{Output format}
Plans are sequences of actions that must end with the action ``done()''. 
Unlike the original setting where the probability of the model is evaluated for each possible action given the observed environment (i.e. pick and place actions for all combinations of objects) , we simply prompt the LLM to output the next action  to accommodate our textual input-output abstraction.
For example: ``\texttt{robot.pick\_and\_place(red block, blue bowl)}\\\texttt{robot.pick\_and\_place(blue block, red bowl)}\\\texttt{done()}''

\subsection{Observation space}
We list all objects with non-zero affordances as observation inputs, e.g. ``\texttt{objects = [red block, blue bowl]}''. Following~\cite{saycan2022arxiv} affordances are implemented with a ViLD object detector~\cite{ALFRED20}.

\subsection{Tasks}
The environment is a tabletop setting in the Pybullet simulator~\cite{coumans2016pybullet} with a UR5 robot and randomly generated  placings of sets of colored blocks and bowls. We specify 10 tasks each with two different object configurations, with varying levels of difficulty and ambiguity.

\renewcommand{\arraystretch}{1.3}

\begin{table}[h!]
\begin{tabularx}{\textwidth}{>{\raggedright\arraybackslash}p{0.48\textwidth}  
                        >{\raggedright\arraybackslash}X}                       
\toprule
\textbf{Task} & \textbf{Objects}\\
\midrule
\rowcolor{white}
\multirow{2}{0.48\textwidth}{Give every bowl an odd number of blocks—or none at all}
  & red/green/blue block, red/blue/green bowls\\
\rowcolor{white}
  & red/blue block, red/blue/green bowls\\[0.3em]

\rowcolor{gray!10}
\multirow{2}{0.48\textwidth}{Put pairs of same-colored blocks into bowls. Leave single blocks.}
  & red/red/blue/green block, red/blue/green bowls\\
\rowcolor{gray!10}
  & red/red/blue/yellow/yellow block, red/blue bowls\\[0.3em]

\rowcolor{white}
Move only the colors that appear exactly once into the yellow bowl,   & red/blue/yellow block, yellow bowl\\
\rowcolor{white}
sort all remaining blocks into bowls that match their color.

  & red block/red block/blue block/blue block/red/blue/green/yellow bowls\\[0.3em]

\rowcolor{gray!10}
\multirow{2}{0.48\textwidth}{Put the blue stuff together\\}
  & blue/red/yellow block, blue/red/yellow bowls\\
\rowcolor{gray!10}
  & blue/blue/red/green block, blue/red/green bowls\\[0.3em]

\rowcolor{white}
\multirow{2}{0.48\textwidth}{Place all blocks on top of their matching bowls\\}
  & red/blue/green block, red/blue/green bowls\\
\rowcolor{white}
  & red/red/blue block, green block/red/blue/green bowls\\[0.3em]

\rowcolor{gray!10}
\multirow{2}{0.48\textwidth}{Separate the blocks by color, putting each color group into a different bowl.}
  & red/blue/green/green block, red/blue/green/yellow bowls\\
\rowcolor{gray!10}
  & red/blue/yellow/yellow block, red/blue bowls\\[0.3em]

\rowcolor{white}
\multirow{2}{0.48\textwidth}{Distribute the blocks as evenly as possible among the available bowls.}
  & red/blue/yellow/red block, red/blue bowls\\
\rowcolor{white}
  & red/blue/green/yellow/red block, red/blue bowls\\[0.3em]

\rowcolor{gray!10}
\multirow{2}{0.48\textwidth}{Stack the blocks that aren't red}
  & red/blue/green/yellow block, blue/green/yellow bowls\\
\rowcolor{gray!10}
  & red/red/green block, blue/green bowls\\[0.3em]

\rowcolor{white}
\multirow{2}{0.48\textwidth}{Put blocks in bowls with non-matching colors\\}
  & red/blue/green/blue/green/yellow bowls\\
\rowcolor{white}
  & red/yellow/green block, blue/green/yellow bowls\\[0.3em]
\rowcolor{gray!10}
\multirow{2}{0.48\textwidth}{Place every block into a bowl, ensuring no block is in a bowl of its own color.}
  & red/blue/green block, red/blue/green/yellow bowls\\
\rowcolor{gray!10}
  & red/blue/green/yellow block, red/blue/green bowls\\
\bottomrule
\end{tabularx}
\caption{Block–bowl manipulation tasks used for the SayCan planner. Bowl and block positions are randomized.}
\label{tab:block_bowl_tasks_clean}
\end{table}

\section{LLM-Planner}
\label{sec:appendix_llm_planner}

LLM-Planner~\cite{song2023llmplanner}: The planner queries an LLM  with the task specification and an object set description of the observed environment, and 9 in-context demonstrations. The demonstrations are selected via a nearest neighbor search using BERT embeddings from a small set of task-plan expert demonstrations (100 samples from training set). The LLM returns a sequence of semantic sub-goals (e.g., \texttt{Navigation fridge}, \texttt{OpenObject fridge}). A separate low-level controller then grounds each sub-goal to primitive actions in the partially observed scene. 

\subsection{Action primitives}
We include seven object interaction actions and a navigation action, as described in~\cite[Sec.\ 3.1]{ALFRED20} and in~\cite[Sec.\ 4.1]{ALFRED20}, respectively. 

\begin{enumerate}[noitemsep,topsep=0pt,left=12pt]
    \item \texttt{PickObject} — pick up (grasp) the target object move it into the agent’s inventory.
    \item \texttt{PutObject} — place the held object at the current focal location or receptacle (e.g., counter top, table).
    \item \texttt{OpenObject} — open an openable object such as a cabinet, drawer, microwave, or door.
    \item \texttt{CloseObject} — close an object previously opened by \texttt{OpenObject}.
    \item \texttt{ToggleOnObject} — switch a toggleable object to its “on’’ state (e.g., faucet, microwave, stove burner).
    \item \texttt{ToggleOffObject} — switch a toggleable object from “on’’ back to “off’’.
    \item \texttt{SliceObject} — use a knife currently held to slice a slice‑able object (e.g., apple, bread, potato) that is specified by the mask.
    \item \texttt{Navigation} — discretised egocentric movement and viewpoint‑control primitives: \texttt{MoveAhead}, \texttt{RotateLeft/Right}, \texttt{LookUp/Down}, which translate or rotate the agent without interacting with objects.
\end{enumerate}

\subsection{Observation space}
The planner recieves a string with visible objects, e.g. ``\texttt{Visible objects: fridge, cabinet, countertop}''.

\subsection{Output format}

Comma-sepparated string of \texttt{Action Object} pairs, e.g.: \textit{``Navigation microwave, OpenObject microwave, PutObject egg microwave,
CloseObject microwave, ToggleObjectOn microwave''}

\subsection{Tasks} We consider tasks from Action Learning From Realistic Environments and Directives (ALFRED)~\cite{ALFRED20}. We evaluate on the seven ALFRED templates:

\begin{itemize}[noitemsep,topsep=0pt,left=5pt]
\item \textit{Clean \& Place} (e.g., ``Wash the mug and put it in the cabinet'') \item \textit{Heat \& Place} (e.g., ``Heat the potato and put it on the counter'')
\item \textit{Cool \& Place} (e.g., ``Cool the soda and put it on the table'')
\item \textit{Examine-in-Light} (e.g., ``Find the keycard and examine it under the lamp'').
\end{itemize}

\textbf{Environments} Experiments are run in the \textit{unseen} split of ALFRED --- i.e., environments differ from the in-context examples from the training set --- comprising eight AI2-THOR~\cite{kolve2017ai2thor} homes with kitchens, bedrooms, bathrooms, and living rooms. The agent must explore cluttered, partially observable rooms containing up to 115 object types.

\section{Scenario generation LLM}
\label{sec:appendix_scenario_gen}

\subsection{SPINE data generator prompt}
\begin{tcolorbox}[colback=gray!3, colframe=black, left=1mm, right=1.5mm, top=1.5mm, bottom=1mm, width=1\textwidth,
coltitle=black,        
colbacktitle=white,    
fonttitle=\bfseries,
boxrule=0.5pt,
boxsep=1mm
] \scriptsize
\begin{minted}{text}
You are generating data for training an llm-based planner, 
like the SPINE paper from ravichandran et al.

Generate a scene graph for training in the following format <OBSERVATION_FORMAT>
for  example, <EXAMPLE_GRAPH>

Make sure all nodes referenced in the conntections are listed in the objects and 
regions list. Provide your answer in the following JSON format:
<OUTPUT_FORMAT>

Add a "description" attribute to each node that provides information.
These will be hidden from the robot

Task generation instructions
- DO NOT reference specific objects or nodes. Make the planner infer theese.
- Tasks should request specific information, not general exploration. 
Make the planner map or inspect certain entities. 
For example, start tasks with phrases such as "what", "I heard", 
"find out", "map", "inspect", "Can I", "is there", and likewise
\end{minted}
\end{tcolorbox}

The variable \textsc{OBSERVATION\_FORMAT} takes the form:
\begin{tcolorbox}[colback=gray!3, colframe=black, left=1mm, right=1.5mm, top=1.5mm, bottom=1mm, width=1\textwidth,
coltitle=black,        
colbacktitle=white,    
fonttitle=\bfseries,
boxrule=0.5pt,
boxsep=1mm
] \scriptsize
\begin{minted}{json}
{"objects": [{"name": "object_1_name", "coords": ["west_east_coordinate", 
    "south_north_coordinate"]}, "..."],
"regions": [{"name": "region_1_name", "coords": ["west_east_coordinate", 
    "south_north_coordinate"]}, "..."],
"object_connections": [["object_name", "region_name"], "..."],
"region_connections": [["some_region_name", "other_region_name"], "..."]
"robot_location": "region_of_robot_location"}
\end{minted}
\end{tcolorbox}

The variable \textsc{EXAMPLE\_GRAPH} takes the form:

\begin{tcolorbox}[colback=gray!3, colframe=black, left=1mm, right=1.5mm, top=1.5mm, bottom=1mm, width=1\textwidth,
coltitle=black,        
colbacktitle=white,    
fonttitle=\bfseries,
boxrule=0.5pt,
boxsep=1mm
] \scriptsize
\begin{minted}{json}
{"objects": [
    {"name": "shed_1", "coords": [78, 9]},
    {"name": "gate_1", "coords": [52, -56]}],
"regions": [
    {"name": "ground_1", "coords": [0, 0]},
    {"name": "road_1", "coords": [5.7, -8.3]},
    {"name": "road_2", "coords": [19.3, -6.5]},
    {"name": "road_3", "coords": [35.7, -12.1]},
    {"name": "road_4", "coords": [52.7, -20]},
    {"name": "road_5", "coords": [57.2, -31.6]},
    {"name": "bridge_1", "coords": [54.3, -46.7]},
    {"name": "road_6", "coords": [52.4, -56.5]},
    {"name": "driveway_1", "coords": [78.4, 9.1]}],
"object_connections": [
    ["shed_1", "driveway_1"],
    ["gate_1", "road_6"]],
"region_connections":[
    ["ground_1", "road_1"],
    ["road_1", "road_2"],
    ["road_2", "road_3"],
    ["road_3", "road_4"],
    ["road_4", "road_5"],
    ["road_5", "bridge_1"],
    ["bridge_1", "road_6"],
    ["road_6", "driveway_1"]],
"robot_location": "ground_1"}
\end{minted}
\end{tcolorbox}

While \texttt{OUTPUT\_FORMAT} comprises a graph, task list, and description.

\subsection{LLM-Planner data generator prompt}

We provide the prompt for the LLM-Planner data generator below.

\begin{tcolorbox}[colback=gray!3, colframe=black, left=1mm, right=1.5mm, top=1.5mm, bottom=1mm, width=1\textwidth,
coltitle=black,        
colbacktitle=white,    
fonttitle=\bfseries,
boxrule=0.5pt,
boxsep=1mm
] \scriptsize
\begin{minted}[breaklines=true]{text}
You are a data generator for synthesizing tasks for the ALFRED simulator.

You should use objects from the following list: 'AlarmClock', 'Apple', 'ArmChair', 'BaseballBat', 'BasketBall', 'Bathtub', 'BathtubBasin', 'Bed', 'Blinds', 'Book', 'Boots', 'Bowl', 'Box', 'Bread', 'ButterKnife', 'Cabinet', 'Candle', 'Cart', 'CD', 'CellPhone', 'Chair', 'Cloth', 'CoffeeMachine', 'CounterTop', 'CreditCard', 'Cup', 'Curtains', 'Desk', 'DeskLamp', 'DishSponge', 'Drawer', 'Dresser', 'Egg', 'FloorLamp', 'Footstool', 'Fork', 'Fridge', 'GarbageCan', 'Glassbottle', 'HandTowel', 'HandTowelHolder', 'HousePlant', 'Kettle', 'KeyChain', 'Knife', 'Ladle', 'Laptop', 'LaundryHamper', 'LaundryHamperLid', 'Lettuce', 'LightSwitch', 'Microwave', 'Mirror', 'Mug', 'Newspaper', 'Ottoman', 'Painting', 'Pan', 'PaperTowel', 'PaperTowelRoll', 'Pen', 'Pencil', 'PepperShaker', 'Pillow', 'Plate', 'Plunger', 'Poster', 'Pot', 'Potato', 'RemoteControl', 'Safe', 'SaltShaker', 'ScrubBrush', 'Shelf', 'ShowerDoor', 'ShowerGlass', 'Sink', 'SinkBasin', 'SoapBar', 'SoapBottle', 'Sofa', 'Spatula', 'Spoon', 'SprayBottle', 'Statue', 'StoveBurner', 'StoveKnob', 'DiningTable', 'CoffeeTable', 'SideTable', 'TeddyBear', 'Television', 'TennisRacket', 'TissueBox', 'Toaster', 'Toilet', 'ToiletPaper', 'ToiletPaperHanger', 'ToiletPaperRoll', 'Tomato', 'Towel', 'TowelHolder', 'TVStand', 'Vase', 'Watch', 'WateringCan', 'Window', 'WineBottle'

The agent has the following capabilities
OpenObject
CloseObject
PickupObject
PutObject
ToggleObjectOn
ToggleObjectOff
SliceObject
Navigation

Your response should be json with the following keys
{tasks: ["description of task 1", "description of task 2", ...],
visible objects: [list of objects in the scene],
reasoning: your reasoning}
\end{minted}
\end{tcolorbox}

\subsection{SayCan data generator prompt}

We provide the prompt for the LLM-Planner data generator below.

\begin{tcolorbox}[colback=gray!3, colframe=black, left=1mm, right=1.5mm, top=1.5mm, bottom=1mm, width=1\textwidth,
coltitle=black,        
colbacktitle=white,    
fonttitle=\bfseries,
boxrule=0.5pt,
boxsep=1mm
] \scriptsize
\begin{minted}[breaklines=true]{text}
You are a task generator for the SayCan pick-and-place environment.
Each task should have a 'raw_input' field describing the goal and a 'config' field specifying 'pick' and 'place' lists. Output a JSON array of task objects.
Examples:
objects = [red block, yellow block, blue block, green bowl]
# put the yellow one the green thing.
robot.pick_and_place(yellow block, green bowl)
done()

objects = [yellow block, blue block, red block]
# move the light colored block to the middle.
robot.pick_and_place(yellow block, middle)
done()

objects = [blue block, green bowl, red block, yellow bowl, green block]
# stack the blocks.
robot.pick_and_place(green block, blue block)
robot.pick_and_place(red block, green block)
done()

objects = [red block, blue block, green bowl, blue bowl, yellow block, green block]
# group the blue objects together.
robot.pick_and_place(blue block, blue bowl)
done()

objects = [green bowl, red block, green block, red bowl, yellow bowl, yellow block]
# sort all the blocks into their matching color bowls.
robot.pick_and_place(green block, green bowl)
robot.pick_and_place(red block, red bowl)
robot.pick_and_place(yellow block, yellow bowl)
done()

Generate 3 tasks using 4 blocks.
\end{minted}
\end{tcolorbox}

\clearpage

\section{Example synthesized environments, tasks, and plans}
\label{sec:appendix_examples}

\subsection{SPINE synthesized data}
\label{sec:appendix_spine_synth_data}
We provide an example synthesized graph and corresponding tasks below. Please note that the output is split to enable readability.
\begin{tcolorbox}[colback=gray!3, colframe=black, left=1mm, right=1.5mm, top=1.5mm, bottom=1mm, width=1\textwidth,
coltitle=black,        
colbacktitle=white,    
fonttitle=\bfseries,
boxrule=0.5pt,
boxsep=1mm
] \scriptsize
\begin{minted}{json}
{ "graph": {
"objects": [
    {"name": "bench_1", "coords": [5, 10],
    "description": "A wooden bench facing the ocean."},
    {"name": "bench_2", "coords": [15, 10],
    "description": "A wooden bench near a lamppost."},
    {"name": "lamppost_1", "coords": [10, 12],
    "description": "A tall lamppost providing light at night."},
    {"name": "food_stall_1", "coords": [20, 8],
    "description": "A stall selling ice cream and snacks."},
    {"name": "food_stall_2", "coords": [25, 8],
    "description": "A stall offering hot dogs and beverages."},
    {"name": "kite_1", "coords": [30, 5],
    "description": "A colorful kite flying high in the sky."},
    {"name": "bicycle_1", "coords": [35, 15],
    "description": "A parked bicycle near the storefronts."},
    {"name": "scooter_1", "coords": [40, 15],
    "description": "A parked electric scooter."},
    {"name": "street_performer_1", "coords": [45, 10],
    "description": "A musician playing guitar."}],
"regions": [
    {"name": "boardwalk_1", "coords": [0, 10],
    "description": "The starting point of the boardwalk."},
    {"name": "boardwalk_2", "coords": [10, 10],
    "description": "A section of the boardwalk with benches."},
    {"name": "boardwalk_3", "coords": [20, 10],
    "description": "A section of the boardwalk with food stalls."},
    {"name": "boardwalk_4", "coords": [30, 10],
    "description": "A section of the boardwalk near kite flyers."},
    {"name": "boardwalk_5", "coords": [40, 10],
    "description": "A section of the boardwalk with street performers."},
    {"name": "beach_1", "coords": [0, 0],
    "description": "A sandy beach area with children playing."},
    {"name": "beach_2", "coords": [10, 0],
    "description": "A beach area with people sunbathing."},
    {"name": "beach_3", "coords": [20, 0],
    "description": "A beach area with kite flyers."},
    {"name": "beach_4", "coords": [30, 0],
    "description": "A beach area with volleyball nets."},
    {"name": "beach_5", "coords": [40, 0],
    "description": "A quieter section of the beach."},
    {"name": "storefront_1", "coords": [0, 20],
    "description": "A row of colorful storefronts."},
    {"name": "storefront_2", "coords": [10, 20],
    "description": "A section with souvenir shops."},
    {"name": "storefront_3", "coords": [20, 20],
    "description": "A section with cafes and restaurants."},
    {"name": "storefront_4", "coords": [30, 20],
    "description": "A section with clothing stores."},
    {"name": "storefront_5", "coords": [40, 20],
    "description": "A section with art galleries."},
    {"name": "parking_lot_1", "coords": [50, 15],
    "description": "A small parking lot for cars."},
    {"name": "palm_tree_1", "coords": [5, 15],
    "description": "A tall palm tree providing shade."},
    {"name": "palm_tree_2", "coords": [25, 15],
    "description": "A palm tree near the food stalls."},
    {"name": "palm_tree_3", "coords": [35, 15],
    "description": "A palm tree near the parked bicycles."},
    {"name": "ocean_1", "coords": [0, -5],
    "description": "The ocean with gentle waves."}],
\end{minted}
\end{tcolorbox}

\begin{tcolorbox}[colback=gray!3, colframe=black, left=1mm, right=1.5mm, top=1.5mm, bottom=1mm, width=1\textwidth,
coltitle=black,        
colbacktitle=white,    
fonttitle=\bfseries,
boxrule=0.5pt,
boxsep=1mm
] \scriptsize
\begin{minted}{json}
"object_connections": [
    ["bench_1", "boardwalk_2"],
    ["bench_2", "boardwalk_2"],
    ["lamppost_1", "boardwalk_2"],
    ["food_stall_1", "boardwalk_3"],
    ["food_stall_2", "boardwalk_3"],
    ["kite_1", "beach_3"],
    ["bicycle_1", "storefront_3"],
    ["scooter_1", "storefront_3"],
    ["street_performer_1", "boardwalk_5"]],
"region_connections": [
    ["boardwalk_1", "boardwalk_2"],
    ["boardwalk_2", "boardwalk_3"],
    ["boardwalk_3", "boardwalk_4"],
    ["boardwalk_4", "boardwalk_5"],
    ["beach_1", "beach_2"],
    ["beach_2", "beach_3"],
    ["beach_3", "beach_4"],
    ["beach_4", "beach_5"],
    ["storefront_1", "storefront_2"],
    ["storefront_2", "storefront_3"],
    ["storefront_3", "storefront_4"],
    ["storefront_4", "storefront_5"],
    ["boardwalk_2", "beach_2"],
    ["boardwalk_3", "beach_3"],
    ["boardwalk_4", "beach_4"],
    ["storefront_3", "parking_lot_1"],
    ["boardwalk_1", "storefront_1"],
    ["boardwalk_5", "storefront_5"],
    ["boardwalk_5", "ocean_1"],
    ["beach_5", "ocean_1"]],
"robot_location": "boardwalk_1"
{"tasks": [
    "What is the closest food stall to the starting point of the boardwalk?",
    "I heard there is a musician playing nearby. Can you find out where?",
    "Map the locations of all benches along the boardwalk.",
    "Inspect the area where kites are flying.",
    "Can I find a place to park a bicycle near the storefronts?",
    "Is there a lamppost providing light near any benches?",
    "Find out which section of the beach has volleyball nets.",
    "Map the palm trees along the boardwalk.",
    "What is the closest section of the boardwalk to the ocean?",
    "Inspect the parking lot for available spaces.",
    "Is there a food stall offering beverages nearby?",
    "Find out where the kite flyers are located.",
    "Map the sections of the boardwalk with street performers.",
    "Is there a quieter section of the beach nearby?",
    "Inspect the area with souvenir shops.",
    "What is the closest palm tree to the food stalls?",
    "Find out where the parked scooters are located.",
    "Map the sections of the boardwalk with lampposts.",
    "Is there a section of the beach with children playing?",
    "Inspect the area with cafes and restaurants.",
    "What is the closest section of the boardwalk to the parking lot?",
    "Find out where the art galleries are located.",
    "Map the sections of the beach with sunbathers.",
    "Is there a section of the boardwalk with cyclists?",
    "Inspect the area with clothing stores."
  ],
"description": "A bustling coastal boardwalk scene flanked by a sandy 
beach on one side and a row of colorful storefronts on the other. The environment 
includes a wooden boardwalk populated with scattered benches, lampposts, 
and occasional food stalls selling snacks and beverages. Close to the boardwalk 
are cyclists and pedestrians enjoying their time, with children playing 
and flying kites on the beach. The ocean waves gently lap against the shore, 
with occasional sailboats in the distance. Behind the storefronts, an array 
of parked bicycles and scooters lines the street, while a small parking 
lot accommodates cars. Tall palm trees provide intermittent shade along 
the boardwalk, and street performers attract crowds. The scene is vibrant 
and dynamic, with seagulls flying above and people capturing the picturesque 
views on their cameras."
}
\end{minted}
\end{tcolorbox}

\subsection{SPINE elicited plan}
We provide the elicited plan with the above semantic graph (\S\ref{sec:appendix_spine_synth_data}) for the task ```Is there a lamppost providing light near any benches?'' is below.
We preface plans with ``// plan'' and observations with ``// observation''.

\begin{tcolorbox}[colback=gray!3, colframe=black, left=1mm, right=1.5mm, top=1.5mm, bottom=1mm, width=1\textwidth,
coltitle=black,        
colbacktitle=white,    
fonttitle=\bfseries,
boxrule=0.5pt,
boxsep=1mm
] \scriptsize
\begin{minted}{text}
// plan
{'primary_goal': 'Determine if there is a lamppost 
providing light near any benches.',
'relevant_graph': 'boardwalk_1, lamppost_1',
'reasoning': 'I cannot reach lamppost_1 from my current location, boardwalk_1. 
I will map boardwalk_1 to try to find a connection.',
'plan': '[map_region(boardwalk_1)]}

// observation
add_nodes({boardwalk_2: {coords: [10.0, 10.0], type: region, description: 
A section of the boardwalk with benches., name: boardwalk_2}}, 
{storefront_1: {coords: [0.0, 20.0], type: region, description: 
A row of colorful storefronts., name: storefront_1}}),
add_connections([boardwalk_1, boardwalk_2], [boardwalk_1, storefront_1]),
update_node_attributes({name: boardwalk_1, description: 
The starting point of the boardwalk.})

// plan
{'primary_goal': 'Determine if there is a lamppost providing light
near any benches.',
'relevant_graph': 'boardwalk_2, lamppost_1, bench_2',
'reasoning': 'I discovered a new region, boardwalk_2, which contains benches. 
I will navigate to boardwalk_2 and explore the region to 
see if there is a lamppost providing light near the benches.',
'plan': '[goto(boardwalk_2), explore_region(boardwalk_2, 3)]'}

// observation
update_robot_location(boardwalk_2)

// plan
{'primary_goal': 'Determine if there is a lamppost providing light near any 
benches.',
'relevant_graph': 'boardwalk_2, lamppost_1, bench_2',
'reasoning': 'I have reached boardwalk_2, which contains benches.
I will now explore the region to see 
if there is a lamppost providing light near the benches.',
'plan': '[explore_region(boardwalk_2, 3)]'}

// observation
add_nodes({bench_1: {coords: [5.0, 10.0], type: object, 
description: A wooden bench facing the ocean., name: bench_1}}, 
{boardwalk_3: {coords: [20.0, 10.0], type: region, 
description: A section of the boardwalk with food stalls., name: boardwalk_3}}),
add_connections([bench_1, boardwalk_2], [bench_2, boardwalk_2], 
[boardwalk_2, lamppost_1], [boardwalk_2, boardwalk_3], 
[beach_2, boardwalk_2]),
update_node_attributes({name: boardwalk_2, 
description: A section of the boardwalk with benches.})'

// plan
{'primary_goal': 'Determine if there is a lamppost 
providing light near any benches.',
'relevant_graph': 'boardwalk_2, lamppost_1, bench_1, bench_2',
'reasoning': 'I discovered that boardwalk_2 has connections to both 
bench_1 and bench_2, as well as lamppost_1. This indicates that lamppost_1
is near the benches and likely provides light to them.',
'plan': '[answer(Yes, there is a lamppost, lamppost_1, 
providing light near the benches, bench_1 and bench_2, on boardwalk_2.)]'}
\end{minted}
\end{tcolorbox}

\subsection{LLM Planner synthesized data}
We provide an example synthesized object set and corresponding tasks below.

\begin{tcolorbox}[colback=gray!3, colframe=black, left=1mm, right=1.5mm, top=1.5mm, bottom=1mm, width=\textwidth,
coltitle=black,        
colbacktitle=white,    
fonttitle=\bfseries,
boxrule=0.5pt,
boxsep=1mm
] 
\begin{minted}[fontsize=\scriptsize, breaklines=true]{json}
{
  "Task description": "Navigate to the DiningTable and use the Knife to slice the bread. Then, place the sliced pieces onto a Plate.",
  "Completed plans": [],
  "Visible objects": [
    "DiningTable",
    "Bread",
    "Knife",
    "Plate",
    "AlarmClock",
    "Desk",
    "Bed",
    "Fridge",
    "Egg",
    "CounterTop",
    "CoffeeMachine"
  ],
  "Next Plans": [
    "Navigation DiningTable",
    "PickupObject Knife",
    "SliceObject Bread",
    "PutObject Knife DiningTable",
    "PickupObject Bread",
    "Navigation Plate",
    "PutObject Bread Plate"
  ]
}
\end{minted}
\end{tcolorbox}

\subsection{LLM Planner elicited plan}

We provide an example elicited plan below.
\begin{tcolorbox}[colback=gray!3, colframe=black, left=1mm, right=1.5mm, top=1.5mm, bottom=1mm, width=1\textwidth,
coltitle=black,        
colbacktitle=white,    
fonttitle=\bfseries,
boxrule=0.5pt,
boxsep=1mm
]  \scriptsize
\begin{minted}{json}
{
    "trial": "trial_T20190908_145356_918528",
    "scene": "FloorPlan10",
    "type": "pick_heat_then_place_in_recep",
    "repeat_idx": 1,
    "goal_instr": "Put a heated apple in the trash can.",
    "initial_high_level_plans": [
      "Navigation apple",
      "PickupObject apple",
      "Navigation microwave",
      "OpenObject microwave",
      "PutObject apple microwave",
      "CloseObject microwave",
      "ToggleObjectOn microwave",
      "ToggleObjectOff microwave",
      "OpenObject microwave",
      "PickupObject apple",
      "CloseObject microwave",
      "Navigation garbagecan",
      "PutObject apple garbagecan"
    ],
    "completed_plans": [
      "Navigation apple",
      "PickupObject apple",
      "Navigation microwave",
      "OpenObject microwave",
      "PutObject apple microwave",
      "CloseObject microwave",
      "ToggleObjectOn microwave",
      "ToggleObjectOff microwave",
      "OpenObject microwave",
      "PickupObject apple",
      "CloseObject microwave",
      "Navigation garbagecan",
      "PutObject apple garbagecan"
    ],
    "failed_plans": [],
    "success": true
  }
\end{minted}
\end{tcolorbox}

\subsection{SayCan synthesized data}

We provide an example synthesized object set below.

\begin{tcolorbox}[colback=gray!3, colframe=black, left=1mm, right=1.5mm, top=1.5mm, bottom=1mm, width=1\textwidth,
coltitle=black,        
colbacktitle=white,    
fonttitle=\bfseries,
boxrule=0.5pt,
boxsep=1mm
]  \scriptsize
\begin{minted}{python}
objects = [red block, green block, blue block, red bowl, green bowl, blue bowl]
# Place all blocks on top of their matching bowls
\end{minted}
\end{tcolorbox}

\subsection{SayCan elicited plan}

We provide an example elicited plan below.

\begin{tcolorbox}[colback=gray!3, colframe=black, left=1mm, right=1.5mm, top=1.5mm, bottom=1mm, width=1\textwidth,
coltitle=black,        
colbacktitle=white,    
fonttitle=\bfseries,
boxrule=0.5pt,
boxsep=1mm
]  \scriptsize
\begin{minted}{python}
objects = [red block, green block, blue block, red bowl, green bowl, blue bowl]
# Place all blocks on top of their matching bowls
robot.pick_and_place(red block, red bowl)
robot.pick_and_place(blue block, blue bowl)
robot.pick_and_place(green block, green bowl)
done()
\end{minted}
\end{tcolorbox}


\end{document}